\pdfoutput=1

\documentclass[11pt]{article}

\usepackage[]{EMNLP2023}

\usepackage{times}
\usepackage{latexsym}
\usepackage{booktabs}
\usepackage{tabularx}

\usepackage[T1]{fontenc}

\usepackage[utf8]{inputenc}

\usepackage{microtype}

\usepackage{inconsolata}

\usepackage{color}

\usepackage{booktabs}

\usepackage{amsmath}
\usepackage{amssymb}
\usepackage[inline,shortlabels]{enumitem}

\usepackage{multirow}
\usepackage[shortlabels,inline]{enumitem}
\usepackage{amsmath} %

\usepackage[colorinlistoftodos,prependcaption,textsize=normalsize]{todonotes}

\usepackage{todonotes}

\usepackage{graphicx}
\graphicspath{ {./figures/} }

\usepackage{tcolorbox}
\usepackage{setspace}

\newcommand\wlcoref{{\texttt{WL-coref}}}

\newcommand\cawcoref{{\texttt{CAW-coref}}}

\usepackage{titlesec}
\titlespacing*{\paragraph}{0pt}{0.5ex}{1ex}

\NewDocumentCommand\emojicaw{}{
    \includegraphics[scale=0.08]{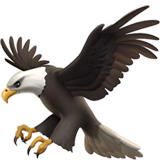}
}

\title{\emojicaw CAW-coref: Conjunction-Aware Word-level Coreference Resolution}

\author{Karel D'Oosterlinck$^{1,*}$, Semere Kiros Bitew$^{1}$, Brandon Papineau$^{2}$ \\ \textbf{Christopher Potts$^{2}$, Thomas Demeester$^{1}$, Chris Develder$^{1}$} \\
  $^1$Ghent University -- imec\qquad$^2$Stanford University  \\
  $^*$\texttt{karel.doosterlinck@ugent.be}
}

\begin{document}

\maketitle

\begin{abstract}
State-of-the-art coreference resolutions systems depend on multiple LLM calls per document and are thus prohibitively expensive for many use cases (e.g., information extraction with large corpora). The leading word-level coreference system (\wlcoref{}) attains 96.6\% of these SOTA systems' performance while being much more efficient. In this work, we identify a routine yet important 
failure case of \wlcoref{}: dealing with conjoined mentions such as \emph{Tom and Mary}. 
We offer a simple yet effective solution that improves the performance on the OntoNotes test set by 0.9\% F1, shrinking the gap between efficient word-level coreference resolution and expensive SOTA approaches by 34.6\%. Our Conjunction-Aware Word-level coreference model (\cawcoref{}) and code is available at \url{https://github.com/KarelDO/wl-coref}.

\end{abstract}

\section{Introduction}

Coreference resolution (or simply \textit{coref}) is the task of clustering mentions in a text, grouping those that refer to the same entity. Coref acts as a fundamental step in many classical NLP pipelines, such as information extraction. Today, however, state-of-the-art (SOTA) coref systems use multiple forward passes of a Large Language model (LLM) \textit{per input document}, making them expensive to train and deploy. This results in limited practical use for classical NLP pipelines, which typically require efficient (and sometimes latency-sensitive) methods.

The most computationally efficient yet competitive neural coref architecture is word-level coref (\wlcoref{}; \citealp{dobrovolskii-2021-word}). 
This method operates by 
\begin{enumerate*}[(i)]
    \item first producing embeddings for each word using one forward pass of a (rather small) LM, then
    \item predicting if pairs of words are coreferent using a lightweight scoring architecture and
    \item finally extracting the 
    spans in the input text associated with these coreferent words.
\end{enumerate*}
Given a text of $n$ words, this incurs a computational complexity of $O(n^2)$, since the method operates on pairs of words. However, SOTA methods typically perform \emph{multiple} forward passes of a (Large) LM per input document, making them unwieldy for many practical applications. Furthermore, these techniques suffer both from high infrastructure costs and latency-issues associated with these large models.

\begin{figure}
    \centering
    \includegraphics[width=\linewidth]{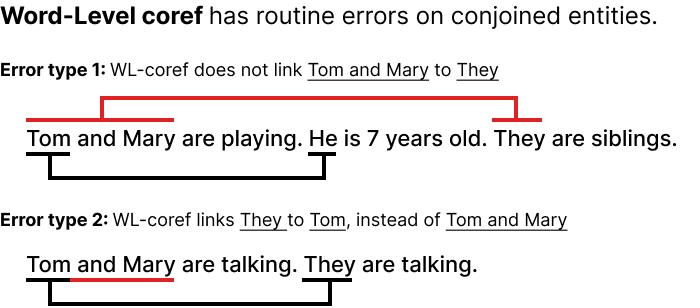}
    \caption{We identify two types of  failure cases for \wlcoref{} when processing conjoined mentions. Our simple solution, \cawcoref{}, addresses these errors. }
    \label{fig:fig1}
\end{figure}

While significantly less complex, \wlcoref{} attains 96.6\% of the performance of the current best coreference model (80.7\% F1 out of 83.3\% F1)\footnote{\citet{dobrovolskii-2021-word} reports a performance of 81.0\% F1 for \wlcoref{} as best performance on the OntoNotes test set. To avoid selecting the best model on the test set, we instead report the test score achieved by our first rerun of \wlcoref{} using their code.}, as measured on the English split of the OntoNotes dataset \citep{pradhan-etal-2012-conll}. What makes this even more impressive is that \wlcoref{} uses one forward pass of a 355M parameter \texttt{roberta-large} encoder \citep{liu2019roberta}, while the state-of-the-art method \citep{bohnet2023coreference} uses multiple forward passes of a 13B parameter \texttt{mT5-XXL} model \citep{xue-etal-2021-mt5}. Thus, \wlcoref{} is the go-to architecture for efficiency-sensitive or long-document coref.

In this work, we describe a fundamental weakness of the \wlcoref{} model in its original formulation, stemming from how the word-level coref step was trained. In particular, starting from a dataset that is annotated at the span-level, a word-level dataset is created by using dependency parsing information to select one head-word per span. This causes ambiguity when mentions are conjoined: two spans representing distinct entities can share the same head-word. For example, the span \emph{Tom and Mary}
is analyzed as containing three entity mentions (\emph{Tom}, \emph{Mary}, and \emph{Tom and Mary}), and both \emph{Tom and Mary} and \emph{Tom} share the same head-word. When the model at inference time tries to refer both to entity \emph{Tom} and entity \emph{Tom and Mary}, two conflicting links to the span \emph{Tom} are predicted. This causes the model to always drop one of the links, degrading performance (Figure~\ref{fig:fig1}).

We resolve this by defining the coordinating conjunction (e.g.~\emph{and}, \emph{or}, \emph{plus}) as head-word when faced with these types of mentions, which is a common approach in linguistics \citep{zoerner1995coordination, progovac1998structure}. Now, the model can learn to systematically link to this conjunction when something is coreferent with \emph{Tom and Mary}, without producing conflicting links. We train a new \wlcoref{} model, called Conjunction-Aware Word-level coreference (\cawcoref{}), and find that this simple fix achieves a significant improvement on the OntoNotes test set: the error difference with the state-of-the-art method shrinks by 34.6\% (i.e.~\cawcoref{} improves the absolute performance of \wlcoref{} from 80.7\% to 81.6\%). Given that this fix incurs no additional model complexity, this gain is an important step forward for efficient coref.

\section{Related Work}
The main competitive approaches to end-to-end coref can be classified into three broad categories: span-based, word-level, and autoregressive coref.

\paragraph{Span-based coreference} \citet{lee-etal-2017-end} introduce \texttt{e2e-coref}, the first end-to-end span-based coref architecture. Starting from word embeddings, the model first predicts which spans are likely a mention. In the second step, coreferent links are predicted between such span-pairs to form coreference clusters. Given a text of $n$ words, this approach incurs $O(n^4)$ computations. Thus, pruning is required to contain the complexity, both for mention prediction and coreference prediction. 

Many follow-up works improved upon this architecture by introducing contextualized embeddings \citep{lee-etal-2018-higher, kantor-globerson-2019-coreference}, LMs for better span representations \citep{joshi-etal-2020-spanbert}, 
ensembling different models for coreference link scoring (\texttt{LingMess}; \citealp{otmazgin2023lingmess}), and distilling the LM backbone for more efficient inference \citep{otmazgin-etal-2022-f}. Still, the theoretical complexity of these approaches remains $O(n^4)$, requiring pruning and leading to poor scaling on long documents.

\paragraph{Word-level coreference} Given an input text, \citet{dobrovolskii-2021-word} proposes to first predict coreference links between words and subsequently extract the spans surrounding words that are found to be coreferent. This lowers the computational cost of the coref architecture to $O(n^2)$. In turn, less aggressive pruning is needed, which resulted in better performance over conventional span-based techniques.\footnote{\texttt{LingMess} \citep{otmazgin2023lingmess} is the only span-based method that outperforms \wlcoref{}, using a lightweight ensembling technique. This technique could be directly applied to \wlcoref{} for potentially a similar performance boost.} \citet{dobrovolskii-2021-word} uses one forward pass of a 355M \texttt{roberta-large} encoder model to form the contextualized word embeddings needed.

\paragraph{Autoregressive coreference} Autoregressive methods iteratively build the coreference structure by running multiple forward passes of an LM backbone. \citet{bohnet2023coreference} introduce a 13B parameter \texttt{mT5-xxl} model called \texttt{link-append}: they run multiple forward passes of the LM over increasingly large chunks of the input text and iteratively predict how to grow the coreference structure. This results in the current state-of-the art model on OntoNotes (+2.6\% F1 over \wlcoref{}). Similarly, \citet{liu2022autoregressive} utilize an 11B parameter \texttt{Flan-T5-xxl} model \citep{chung2022scaling} and predict a sequence of structure-building actions when regressing over the input text (\texttt{ASP}). \citet{wu2020corefqa} introduce \texttt{corefqa}, formulating coref as a series of question-answering tasks, run multiple forward passes of an LM to build the coreference structure and use extra QA data for augmentation.

In general, the autoregressive methods outperform span-based and word-level coreference, but at great computational cost. All these methods require at least $O(n)$ forward passes of an LM per input document, while span-based or word-level techniques require only one. While some of these computations could be parallelized, running $O(n)$ LM forward passed \emph{per input document} is exceedingly expensive.

Additionally, the \texttt{mT5-xxl} and \texttt{T0} models used by SOTA methods contain many more parameters compared to the \texttt{roberta-large} model used by \wlcoref{} (13B and 11B respectively, compared to 355M), making these models less accessible to train and deploy. \citet{liu2022autoregressive} show that when using an LM comparable in size to the one used by \wlcoref{}, their performance using autoregressive coreference is actually worse. Thus, word-level coreference is the most efficient method in terms of memory requirements and computational scaling.

\paragraph{Error analysis of coreference models} 
\citet{porada2023investigating} investigate types of errors in recent coref models, including \wlcoref{}. Based on the hypothesis that distinct datasets operationalize the task of coreference differently, they perform generalization experiments between multiple datasets and analyze different types of model error. One of their findings suggests that coref for nested mentions is still hard in general.

In this work, we highlight a failure case of \wlcoref{}, namely, coreference with conjoined entities (i.e.~coordinated noun phrases). We propose and empirically validate a simple yet effective solution.

\section{The \wlcoref{} model}
We briefly summarize the architecture used by \citet{dobrovolskii-2021-word} and refer to the original publication for a full overview.

\paragraph{Step~1 -- Word Representations:}First, contextualized word representations are created using one forward pass of an LM backbone and a learned averaging over constituent toMarys.

\paragraph{Step~2 -- Word-Level Coreference:} To create word-level links, a first \emph{coarse antecedent scoring} is constructed between all pairs of words using a learned bilinear function.

For each word, the top $k$ coarse antecedents are considered in a \emph{fine antecedent scoring step}, using a trained feed forward neural network.
The final antecedent scores are given by the sum of the coarse and fine scores. These antecedent scores between pairs of words are used to infer the most likely word-level coreference clustering.
The words found to be part of a coreference cluster are passed on to Step~3.

\paragraph{Step~3 -- Span Extraction:} For each coreferent word, the mention span surrounding it is extracted. This is done using a small feed-forward neural network applied to the contextualized word embeddings, followed by a  convolutional layer which predicts probabilities for start and end span boundaries. This step is applied individually for each coreferent word and thus is not directly aware of the global clustering produced in Step~2.

\paragraph{Creating word-level data:} To train both steps, \citet{dobrovolskii-2021-word} uses syntactic information to decompose the span-based OntoNotes dataset into a word-level version and a word-to-span dataset. 

The crucial step in this decomposition is selecting one head-word per span. Clearly, these head-words need to be as representative as possible of the entity mentioned in the span, so as to allow the word-level linking to perform well. Additionally, the head-words should be systematically picked so that the span extraction step has an easy time learning to extract the correct span surrounding a coreferent head-word.

\citet{dobrovolskii-2021-word} picks head-words using dependency parsing information already present in the OntoNotes dataset. Given a span, the method selects the head-word as the word in the span which depends on a word outside of the span. If none or multiple of such words are found, the right-most word of the span is selected as head-word.

\section{Failure Modes of \wlcoref{}}
We describe the two failures cases of \wlcoref{} outlined in Figure~\ref{fig:fig1} and propose a simple solution.

\begin{table*}[]\centering
\begin{footnotesize}
\begin{tabularx}{\textwidth}{@{}l|ll|l|ccc|ccc|ccc|c@{}}
\toprule
 & \multicolumn{2}{c}{\textbf{LM}} & \multicolumn{1}{c}{\textbf{Link}}& \multicolumn{3}{c}{\textbf{MUC}} & \multicolumn{3}{c}{\textbf{B\textsuperscript{3}}} & \multicolumn{3}{c}{\textbf{CEAF\textsubscript{$\phi{}4$}}} & \textbf{Avg.}           \\ 
 & calls & params.& compl. & P      & R     & F1     & P     & R     & F1    & P      & R      & F1     & \textbf{F1} \\ \midrule

\texttt{link-append} & $O(n)$ & 13B & / & \textbf{87.4}            & 88.3        & \textbf{87.8}      & 81.8        & \textbf{83.4}       & \textbf{82.6}       & 79.1       & \textbf{79.9}       & \textbf{79.5}        & \textbf{83.3}                 \\ 
\texttt{corefqa} & $O(n^2)$ & 340M & / & 88.6 & 87.4 & 88.0 & \textbf{82.4} & 82.0 & 82.2 & \textbf{79.9} & 78.3 & 79.1 & 83.1 \\
\texttt{ASP} & $O(n)$ & 11B & / & 86.1            & \textbf{88.4}        & 87.2      & 80.2        & 83.2       & 81.7       & 78.9      & 78.3       & 78.6        & 82.5               \\ \midrule
\texttt{LingMess}     & 1& 355M      & $O(n^4)$      & 85.1    & 88.1      & \textbf{86.6}        & \textbf{78.3}       & \textbf{82.7}       & \textbf{80.5}      & 76.1       & 78.5       & 77.3        & 81.4        \\ 
\texttt{s2e}       & 1 & 355M   & $O(n^4)$         & \textbf{85.2} & 86.6       & 85.9        & 77.9       & 80.3      & 79.1       & 75.4        & 76.8        & 76.1        & 80.3        \\ 
\texttt{CAW} (ours) &1 & 355M & $O(n^2)$           & 85.1       & \textbf{88.2}        & \textbf{86.6}       & 77.0      & 78.0       & 77.5       & \textbf{78.0}        & \textbf{83.2}        & \textbf{80.6}        & \textbf{81.6}         \\ 
\texttt{WL}\textsuperscript{\textdagger}  &1 & 355M& $O(n^2)$           & 84.8       & 87.5      & 86.1       & 76.1      & 76.7      & 76.6      & 77.1       & 82.1       & 79.5       & 80.7        \\ \bottomrule

\end{tabularx}
\caption{Results on the OntoNotes 5.0 English test set. Scores calculated with official scorer \citep{pradhan-etal-2014-scoring} or taken from original publication if available. \textbf{Avg. F1} is the main metric. We report the amount of LM calls and parameters of the LM used, as well as the coreference linking complexity if applicable. \textdagger~ \citet{dobrovolskii-2021-word} reports an Avg. F1 of 81.0 as the best \wlcoref{} run on the test set, while we report the result of our first run for both \wlcoref{} and \cawcoref{}. }
\label{table:results}
\end{footnotesize}
\end{table*}

\paragraph{Entity Conjunction:} \wlcoref{} is unable to fully solve routine examples where the conjunction of two or more mentions (e.g.~via the use of the coordinating conjunction \emph{and}) forms a new mention in the discourse. Consider the first example from Figure \ref{fig:fig1}: \emph{Tom and Mary are playing. He is 7 years old. They are siblings}. Following how head-words were defined in \citealt{dobrovolskii-2021-word}, both the head-word for the mention \emph{Tom and Mary} and the mention \emph{Tom} coincide. At inference time, the word-level coreference step will thus predict both the coreferent links \emph{Tom} -- \emph{He} and \emph{Tom} -- \emph{They}. Since the model does not predict a link \emph{He} -- \emph{They}, one of these two predicted links must be dropped in order to arrive at a consistent clustering. Thus, the model is unable to correctly output both coreferent clusters in this trivial example.

\paragraph{Nested Span Extraction:} Given a coreferent head-word, \wlcoref{} sometimes struggles to extract the correct span boundaries surrounding this head-word when multiple valid options are possible. Consider the second example from Figure \ref{fig:fig1}: \emph{Tom and Mary are talking. They are talking}. \wlcoref{} correctly predicts the word-level link between \emph{Tom} -- \emph{They}, but fails to extract the span \emph{Tom and Mary} in the subsequent step. This is most likely caused by the span extraction step operating independently on every coreferent head-word: no explicit information about the \emph{Tom} -- \emph{They} link is taMary into account when deciding between \emph{Tom} and \emph{Tom and Mary}, and this decision is thus ambiguous.

\paragraph{Proposed Solution:} Both failure modes are rooted in the same fundamental problem: there is no unique one-to-one relation between head-words and spans. This causes  issues both when predicting word-level links and when performing span extraction, specifically when dealing with nesting.

We propose to solve this by changing how head-words are defined on conjoined mentions. When creating the word-level training data, we use part-of-speech tags supplied in the OntoNotes dataset to detect if a coordinating conjunction (e.g.~\emph{and}, \emph{or}, \emph{plus}) is present in a span. Then we check the relative depth of the conjunction in the dependency parse of the span. If it is less than two steps away from the head-word of the span, it is selected as new head-word. This selects \emph{and} as head-word in the span \emph{Tom and Ann}, but not in the span \emph{David, whose children are called Tom and Ann}. Thus, we have defined a systematic way of picking head-words for conjoined mentions, in a way that they do not conflict with any of the head-words for the nested mentions.

\section{Experiments and Results}
We use our new word-level dataset to train \cawcoref{}, a new instance of the \wlcoref{} architecture. Using our altered notion of head-words, we train and evaluate this model on the English OntoNotes dataset without changing any hyperparameters compared to the default \wlcoref{} run. We immediately find an absolute performance increase of 0.9\% F1, setting the performance of \cawcoref{} at 81.6\% F1. This shrinks the relative gap between efficient coref and expensive SOTA approaches by 34.6\%, which is certainly not trivial since gains on OntoNotes have been hard to come by in recent years. 

The full breakdown of the results in function of the official evaluation metrics \citep{vilain-etal-1995-model, Bagga1998AlgorithmsFS, luo-2005-coreference, pradhan-etal-2012-conll}
is given in Table~\ref{table:results}. \cawcoref{} even outperforms \texttt{LingMess}, the best span-based method, which uses ensembling to achieve a significant performance boost. Potentially, such an ensembling technique could be applied to further boost \cawcoref{} performance as well.

In total, we found that 1.17\% of spans across the English OntoNotes train and development split were such conjoined entities. Supplementary to our empirical analysis, we show the qualitative improvement of \cawcoref{} on a list of simple examples in Appendix \ref{app:qual-examples}.

\section{Conclusion}

Neural coreference resolution techniques should be efficient in order to maximize real-word impact. In this work, we outlined two failure cases of the efficient word-level coreference resolution architecture and addressed them with one simple fix. Our new model, Conjunction-Aware Word-level coreference (\cawcoref{}), shrinks the performance gap between efficient and state-of-the-art coreference by 34.6\%, and is currently the most performant efficient neural coreference model.

\clearpage

\clearpage

\section*{Limitations}
There are always more distinct spans than words in a text, thus it is not always possible to uniquely pick a head-word per span. For example, our proposed solution can't fully handle sequential conjunctions such as \emph{Tom and Mary and David}, since this span contains only 5 words but 6 mentions: \emph{Tom}, \emph{Tom and Mary}, \emph{Mary}, \emph{Mary and David}, \emph{David}, and \emph{Tom and Mary and David}. Luckily, we did not observe any such dense references in the dataset.

Our procedure of selecting a new head-word for conjunctions relies on syntactic information in the form of part-of-speech tags and dependency parses. OntoNotes features several instances where conjunctions are formed using commas or hyphens, such as in the span \emph{Tom~,~Mary} or \emph{Tom~-~Mary}. Here, the comma and hyphen should take on the role as head-word of the conjunction, but this is much harder to detect using the syntactic information present.

Future work could focus on resolving both these issues to further boost the performance of efficient Conjunction-Aware Word-level coreference resolution.

\section*{Acknowledgements}
We are grateful to our anonymous reviewers for their meticulous reading and valuable comments. Karel D'Oosterlinck is funded by an FWO Fundamental Research PhD Fellowship (11632223N).

\bibliography{anthology,custom}
\bibliographystyle{acl_natbib}

\clearpage
\onecolumn

\appendix

\section{Qualitative Examples}
\label{app:qual-examples}

\begin{table}[t]\centering
    \begin{tabularx}{\textwidth}{l|l|X|l} \toprule
        \textbf{Model} & \textbf{Step} & \textbf{Prediction} & \textbf{Correct} \\ \midrule
         \wlcoref{} & word &  \colorbox{cyan}{Tom} and Anna are talking. \colorbox{cyan}{They} are talking. & \textcolor{teal}{Yes}\\
         \wlcoref{} & span & \colorbox{cyan}{Tom} and Anna are talking. \colorbox{cyan}{They} are talking. & \textcolor{red}{No}\\ 
         \cawcoref{} & word &  \colorbox{cyan}{Tom} and Anna are talking. \colorbox{cyan}{They} are talking. & \textcolor{teal}{Yes}\\
         \cawcoref{} & span & \colorbox{cyan}{Tom and Anna} are talking. \colorbox{cyan}{They} are talking. & \textcolor{teal}{Yes}\\ \midrule
         \wlcoref{} & word &  \colorbox{cyan}{My} friend David and \colorbox{cyan}{my} dad Bert are talking. They are talking. & \textcolor{red}{No}\\
         \wlcoref{} & span & \colorbox{cyan}{My} friend David and \colorbox{cyan}{my} dad Bert are talking. They are talking. & \textcolor{red}{No}\\ 
         \cawcoref{} & word &  \colorbox{cyan}{My} friend David \colorbox{magenta}{and} \colorbox{cyan}{my} dad Bert are talking. \colorbox{magenta}{They} are talking. & \textcolor{teal}{Yes}\\
         \cawcoref{} & span & \colorbox{magenta}{\colorbox{cyan}{My} friend David and \colorbox{cyan}{my} dad Bert} are talking. \colorbox{magenta}{They} are talking. & \textcolor{teal}{Yes}\\ \midrule
         \wlcoref{} & word &  The Guardian and The Chronicle had a secret meeting . Both newspapers are on thin ice . & \textcolor{red}{No}\\
         \wlcoref{} & span & The Guardian and The Chronicle had a secret meeting . Both newspapers are on thin ice . & \textcolor{red}{No}\\ 
         \cawcoref{} & word &  The Guardian \colorbox{cyan}{and} The Chronicle had a secret meeting . Both \colorbox{cyan}{newspapers} are on thin ice . & \textcolor{teal}{Yes}\\
         \cawcoref{} & span & \colorbox{cyan}{The Guardian and The Chronicle} had a secret meeting . \colorbox{cyan}{Both newspapers} are on thin ice . & \textcolor{teal}{Yes}\\ \bottomrule
    \end{tabularx}
\caption{Three hand-crafted examples and their word-level and span-level predictions for \wlcoref{} and \cawcoref{}. Coreferent predictions are indicated with a colored box, where each unique entity has the same color. Predictions are considered \textcolor{teal}{correct} or \textcolor{red}{not correct} for their respective step in the word-level pipeline.}
\label{table:app-examples}
\end{table}

Three qualitative examples comparing \wlcoref{} and \cawcoref{} with the word-level and span-level predictions are given in Table \ref{table:app-examples}.

\end{document}